\documentclass{article}
\usepackage{arxiv}

\usepackage[T1]{fontenc}    
\usepackage[utf8]{inputenc} 
\usepackage{hyperref}       
\usepackage{url}            
\usepackage{booktabs}       
\usepackage{amsfonts}       
\usepackage{nicefrac}       
\usepackage{microtype}      
\usepackage{multirow}
\usepackage{graphicx}
\usepackage{natbib}
\usepackage{doi}
\usepackage{subcaption} 
\usepackage{float}   
\usepackage{amssymb}
\usepackage{amsmath}
\usepackage{mathtools}
\usepackage{latexsym}

\usepackage{tikz}
\usetikzlibrary{arrows.meta, positioning, calc, shapes, shadows}

\definecolor{darkblue}{rgb}{0, 0, 0.5}
\hypersetup{colorlinks=true, citecolor=darkblue, linkcolor=darkblue, urlcolor=darkblue}

\usepackage[table]{xcolor}
\usepackage{tabularray}

\usepackage[main=english, russian]{babel}
\usepackage{CJKutf8}

\normalfont
\DeclareFontFamilySubstitution{T2A}{ppl}{cmr}

\title{Probing the Geometry of Color Representations in Multilingual Language Models}

\date{} 					

\author{ 
    Pablo Loyola \\
	Rakuten Group, Inc.\\    
	\texttt{pablo.a.loyola@rakuten.com} \\    
    \And
    Edison Marrese-Taylor \\
    Graduate School of Engineering, The University of Tokyo, Japan\\
	\texttt{emarrese@weblab.t.u-tokyo.ac.jp} \\
	\And
    {Andrés Hoyos-Idrobo} \\
	Rakuten Group, Inc.\\
	\texttt{andres.hoyosidrobo@rakuten.com} \\
}

\renewcommand{\headeright}{}
\renewcommand{\undertitle}{}
\renewcommand{\shorttitle}{Probing Color Geometry in Multilingual LMs}

\newcommand{\pipelinePalette}{2}

\hypersetup{
pdftitle={Probing the Geometry of Color Representations in Multilingual Language Models},
pdfsubject={cs.CL, cs.LG},
pdfauthor={Pablo Loyola, Edison Marrese-Taylor, Andr\'es Hoyos-Idrobo},
pdfkeywords={perceptual grounding, multilingual language models,  color representations},
}

\begin{document}
\maketitle

\begin{abstract}
Do multilingual language models learn a universal internal geometry for color, and is it perceptually grounded or primarily linguistic? We decompose color representations into two geometric axes: \emph{(P1)} alignment to perceptual color space, and \emph{(P2)} cross-lingual symmetry. Analyzing layerwise embeddings across models and multilingual datasets, we find that perceptual grounding is weak: alignment to CIE--LAB peaks in intermediate layers, then collapses toward output layers. Cross-lingual symmetry degrades with depth, suggesting later layers encode language-specific structure. Functionally, we show that geometric properties predict task performance: models achieve 96.7\% accuracy on color classification at layers where \emph{P1} alignment peaks, with \emph{P1} alignment ($r=0.70$) significantly correlating with downstream capabilities. Our results reveal a dissociation: text-only LMs learn universal color-\emph{language} geometry, not universal color-\emph{perception} geometry, providing diagnostics for where perceptual structure is preserved in language models.
\end{abstract}


\section{Introduction}
Do Language Models (LMs) build latent geometric structures that systematically organize meaning? Recent work suggests neural representations are best viewed as geometric objects---manifolds or trajectories in activation space. In language models, latent factors trace continuous curves in hidden states~\citep{dohmatob2023origins,engels2025not}, while in vision, object manifolds become more compact across layers~\citep{ansuini2019intrinsic,cohen2020separability}. Competing hypotheses have emerged: the Linear Representation Hypothesis~\citep{park2024linear} posits that concepts correspond to linear directions, while the Platonic Representation Hypothesis~\citep{huh2024platonic} argues that representations converge toward shared structure across architectures. Yet how these geometries align with real-world perceptual structure remains unclear.

We use color as a concrete testbed. It has well-studied perceptual geometry (CIE--LAB) as a proxy for human similarity, and multilingual datasets exposing systematic cross-linguistic variation~\citep{winawer2007russian,gibson2017color}. This makes it ideal for disentangling perceptual structure from linguistic categorization, and for testing whether LMs learn compact, perceptually grounded geometry shared across languages.

We ask: \emph{What geometric structure do multilingual language models learn for color---one aligned with perceptual space (CIE--LAB), language-driven categorical geometry, or a mix that shifts across layers?} This question exposes two geometric axes: \emph{(P1)} perceptual alignment---does similarity geometry match CIE--LAB, and at which layers? \emph{(P2)} cross-lingual symmetry---is geometry shared across languages or does it drift with linguistic variation?

Our approach is pragmatic: we do not expect text-only LMs to reproduce human perception. Instead, we use LAB as a template to measure to what extent color structure is preserved or altered at each layer. We analyze layerwise representations across models and multilingual datasets, assessing kernel alignment with CIE--LAB and measuring cross-lingual drift.

We find a clear dissociation: text-only LMs learn a compact color-language geometry that is partly universal\footnote{Throughout, ``universal'' refers to trends consistent across diverse text-only causal LMs. Multimodal or vision-conditioned models can deviate due to different training objectives, which we treat as informative evidence about how training signals shape representational geometry.} across languages, yet perceptual (LAB) alignment is weak and, when present, peaks in intermediate layers before collapsing at output. Cross-lingual congruence is highest in early layers and degrades with depth. Functionally, these geometric properties predict downstream task performance, with models achieving 96.7\% accuracy on color classification at layers where alignment peaks.

\paragraph{Contributions.}
We introduce internal color geometry as a testbed for analyzing how multilingual LMs organize perceptual-like domains, develop a layerwise framework combining perceptual kernel alignment and cross-lingual symmetry analysis, and provide evidence that LMs learn stable color-language geometry without strong perceptual grounding. This reveals that universality can emerge without perceptual alignment, with functional implications for layer selection in downstream applications.

\subsection{Problem Formulation: Internal Color Geometry}
\label{sec:problem}

We analyze the geometry of color meaning in multilingual LMs. Given colors with multilingual descriptions, we ask what geometric patterns the model expresses and how they relate to perceptual color space (CIE--LAB) and cross-lingual invariance. We use perceptual color spaces as reference geometries—not because text-only models should match human vision, but to measure what geometry language supervision produces and where it aligns with perception.

We focus on two geometric properties of layerwise representations. First, \textbf{(P1) perceptual alignment}: does the model's similarity geometry match perceptual space, and at which layers? We test whether neighborhoods or kernels from the model distance matrix match those from CIE--LAB, up to global distortions. Second, \textbf{(P2) cross-lingual symmetry}: given colors described in multiple languages, is geometry stable or does linguistic form distort pairwise structure? We quantify this by comparing language-specific kernels on matched samples.

\section{Related Work}
\label{sec:related-work}

\paragraph{Geometric methods for analyzing representations.}
Representations in deep networks often lie on low-dimensional manifolds~\citep{ansuini2019intrinsic,cohen2020separability,modell2025origins}, compared via RSA~\citep{kriegeskorte2008representational}, CKA~\citep{kornblith2019similarity}, and kernel methods~\citep{gretton2005measuring}. We extend this toolkit to a concrete grounded domain, i.e., color, where ground-truth perceptual structure (CIE--LAB) exists, enabling direct tests of whether learned geometry reflects perception or language-specific categorization. Crucially, we show that low dimensionality need not imply perceptual grounding.

\paragraph{Representation structure in language models.}
The Platonic Representation Hypothesis~\citep{huh2024platonic} posits convergence to shared representations across models and modalities, while work on multilingual models shows cross-lingual similarity strongest in early layers, diverging with depth~\citep{conneau2017word,libovicky2020language,chi2020finding}. 
Layerwise analyses also show that intermediate layers often carry more useful representations than the final one~\citep{skean2025layer}, consistent with the perceptual window we find for color.
We provide the first systematic analysis decomposing color geometry into perceptual alignment \textbf{(P1)} and cross-lingual symmetry \textbf{(P2)}, revealing that universality emerges without perceptual grounding. This challenges assumptions that universal representations imply perceptual grounding.

\paragraph{Color as a testbed for grounding.}
Color naming has been central in linguistics~\citep{berlin1991basic,regier2007color}, and recent work examines LLM color-naming behavior through output accuracy\citep{abdou2021can, loyola2023perceptual, winn2018lighter}. 
We focus instead on internal geometry across layers: testing whether color representations are compact, perceptually aligned, and language-invariant. 
This geometric lens isolates where and why perceptual grounding fails in text-only models, revealing that perceptual structure is transiently expressed in intermediate layers but collapses toward output---a finding with direct implications for layer selection in applications.

\section{Methods}
\label{sec:methods}
This section specifies how we extract representations from language models, 
define the induced geometry, and compute the two axes of analysis (\emph{P1}--\emph{P2}) 
together with uncertainty estimates. Our approach combines \emph{geometric analysis} 
of internal representations with \emph{functional probing} of task performance. 
We directly analyze the geometric structure of representations to understand what 
color-related knowledge is encoded and how it evolves across layers, then validate 
whether geometric quality predicts downstream task utility. 
This reveals both \emph{where} (at which layers) and \emph{how} (through what geometric properties) 
models organize color meaning, and whether this organization matters functionally.

\begin{figure}[p]
\centering
\input{content/figure_pipeline_v1}
\end{figure}

\subsection{Notation and Representations}
\label{subsec:notation}
Let $\mathcal{D} = \{(\mathbf{c}_i, \mathbf{d}_i, \ell_i)\}_{i=1}^{N}$ denote a dataset of $N$~color descriptions, where $\mathbf{c}_i \in \mathbb{R}^3$ is the CIE--LAB coordinate, $\mathbf{d}_i$ is the tokenized description, and $\ell_i \in \mathcal{L}$ is a language label. For a transformer with layers $l\in\{0,\dots,L\}$, we extract sequence-level representations $\mathbf{z}_i^{(l)}\in\mathbb{R}^{d_l}$ via pooling (mean or max) over token-level hidden states, where $d_l$ is the hidden dimension at layer~$l$. We denote the full representation matrix as $\mathbf{Z}^{(l)}=[\mathbf{z}_1^{(l)},\dots,\mathbf{z}_N^{(l)}]^\top\in\mathbb{R}^{N\times d_l}$.

We use cosine distance $d_{\cos}(\mathbf{u},\,\mathbf{v}) = 1-\frac{\langle \mathbf{u},\,\mathbf{v}\rangle}{\|\mathbf{u}\|\|\mathbf{v}\|}$ (appropriate for LayerNorm’ed representations) and Euclidean distance\footnote{When vectors are approximately normalized, $d_2^2(\mathbf{u},\,\mathbf{v}) \approx 2\,d_{\cos}(\mathbf{u},\,\mathbf{v})$.}, $d^2(\mathbf{u},\,\mathbf{v})=\|\mathbf{u}-\mathbf{v}\|^2$, inducing a distance matrix $\mathbf{D}_{\mathcal{M}}^{(l)} \in \mathbb{R}_{+}^{N\times N}$ with entries $\mathbf{D}_{\mathcal{M}}^{(l)}(i,j) = d_{\mathcal{M}}\big(\mathbf{z}_i^{(l)}, \,\mathbf{z}_j^{(l)}\big)$. 
For perceptual reference, we use $\Delta E$ distance~\citep{sharma2005ciede2000} in CIE--LAB, $\mathbf{D}_{\mathrm{LAB}} \in \mathbb{R}_{+}^{N \times N}$ with entries $\mathbf{D}_{\mathrm{LAB}}(i,j) = d_{\mathrm{LAB}}(\mathbf{c}_i,\,\mathbf{c}_j)$. 
We adopt LAB as an operational proxy for perceptual similarity, not claiming text-only models should reproduce human vision. 
The LAB distance offers a well-calibrated perceptual baseline since a difference of $\Delta E \approx 2.3$ corresponds to a just-noticeable difference under controlled settings~\citep{sharma2005ciede2000}, making it an interpretable reference for human color discrimination.

For kernel-based comparisons, we form kernels via RBF, $\mathbf{\widetilde{K}}_{\mathcal{T}}^{(l)}(i,j) = \exp\bigl(-\mathbf{D}_{\mathcal{S}}^{(l)}(i,j)^2 / 2\sigma^2\bigr)$, where $\mathcal{T} = \{\mathrm{LAB},\,\mathcal{M}\}$, with $\sigma$ set by median pairwise distance. We use centered and normalized kernels:
\begin{equation}
\overline{\mathbf{K}} = \mathbf{H}\,\mathbf{\widetilde{K}}\,\mathbf{H} \, \text{(centered)}, \quad \mathbf{H} = \mathbf{I} - \tfrac{1}{N}\mathbf{1}\mathbf{1}^\top, \quad \text{and} \quad
\mathbf{K} = \frac{\overline{\mathbf{K}}}{\|\overline{\mathbf{K}}\|_\textsc{F}} \, \text{(centered and normalized)},
\label{eq:kernel_centering}
\end{equation}
where $\|\cdot\|_\textsc{F}$ is the Frobenius norm. Centering removes global shift effects; normalization bounds alignment measures, enabling direct comparison across layers and datasets. The median heuristic for $\sigma$ adapts to the characteristic scale of each representation space, avoiding manual tuning.

\subsection{P1: Perceptual Alignment}
\label{subsec:p1}
We quantify how closely the pairwise similarity structure in model space~$\mathbf{K}_{\mathcal{M}}^{(l)}$ matches perceptual similarity~$\mathbf{K}_{\mathrm{LAB}}$ at each layer~$l$ via Centered Kernel Alignment (CKA)~\citep{cristianini2001kernel,cortes2012algorithms}.
CKA is invariant to orthogonal transformations and isotropic scaling, making it robust to representational choices that preserve relational structure, such as rotation or uniform rescaling. Unlike pointwise metrics, CKA compares \emph{entire similarity matrices} and captures whether the model preserves the global pattern of color similarity. A high CKA score indicates that if two colors are perceptually close in LAB space, they are also close in model space, and vice versa—even if their absolute distances differ. This makes CKA well-suited for evaluating whether language models learn perceptual-like \emph{geometry} (i.e., which items are close to each other) rather than exact metric properties (i.e., precise distances). Kernel alignment is defined as follows, 
\begin{equation}
\mathrm{KA}\!\left(\mathbf{K}_{\mathcal{M}}^{(l)}, \,\mathbf{K}_{\textsc{lab}}\right)
=
\left\langle {\mathbf{K}}_{\mathcal{M}}^{(l)}, \, {\mathbf{K}}_{\textsc{lab}}\right\rangle_\textsc{F},
\label{eq:ka}
\end{equation}
where $\langle\cdot,\,\cdot\rangle_\textsc{F}$ is the Frobenius inner product and
${\mathbf{K}}$ denotes the centered and normalized kernel, Equation~\ref{eq:kernel_centering}.
We also report Spearman correlation between the upper-triangular entries of
$\mathbf{D}_{\mathcal{M}}^{(l)}$ and $\mathbf{D}_{\mathrm{LAB}}$.

\subsection{P2: Cross-Lingual Symmetry}
\label{subsec:p2}
We assess whether different languages induce the same internal geometry over a shared set of colors at each layer~$l$ via congruence and drift on matched color sets.

Let $\mathcal{S}$ be a set of $n$ matched color items described in languages $\ell$ and $\ell'$. 
We compute their representations, $\mathbf{Z}^{(l)}_{\ell} \in \mathbb{R}^{n \times d_l}$ and $\mathbf{Z}^{(l)}_{\ell'} \in \mathbb{R}^{n \times d_l}$. Then, we build kernels $\mathbf{K}^{(l)}_{\ell} \in \mathbb{R}^{n \times n}$ and $\mathbf{K}^{(l)}_{\ell'} \in \mathbb{R}^{n \times n}$ using the same distance function, and compute all symmetry metrics on these matched, consistently ordered sets.
In particular, we quantify symmetry via 
\begin{equation}
\mathrm{Cong}^{(l)}(\ell,\,\ell') = \mathrm{KA}(\mathbf{K}^{(l)}_{\ell}, \,\mathbf{K}^{(l)}_{\ell'}) \quad \text{and} \quad \mathrm{Drift}^{(l)}(\ell, \,\ell') = \| {\mathbf{K}}^{(l)}_{\ell} - {\mathbf{K}}^{(l)}_{\ell'}\|_\textsc{F}    
\end{equation} 
Congruence measures global geometric agreement; drift provides geometric change. Thus, high congruence indicates that the model treats colors similarly across languages. For instance, if ``red'' and ``orange'' are close in English representations, their equivalents in other languages are also close by the same relative amount. Thus, it tests whether multilingual LMs learn a shared, language-invariant color geometry or if linguistic form systematically distorts pairwise structure.

\section{Data}
\label{sec:data}
We use two complementary datasets with human-produced color descriptions: \textsc{MLMC} (pragmatic, e.g., \textit{dark green}) and \textsc{M-ColorNames} (figurative, e.g., \textit{garden's darkest night}). This contrast captures different modes of color language use. We also use \textsc{Munsell} chips as perceptual reference.

\paragraph{Many Languages, Many Colors (\textsc{MLMC}) dataset.}
\textsc{MLMC}~\citep{kim2019color} comes from a controlled multilingual color-naming study, with \emph{pragmatic} descriptions (e.g., \textit{dark green}, \textit{light orange}). We use a filtered version:
(i) we remove invalid entries: empty or non-word strings;
(ii) we keep languages with at least 30 valid examples, leaving ten languages total.
All entries have LAB coordinates.
The dataset achieves 100\% hue coverage (measured using 36 evenly spaced 10\textdegree{} bins) and spans $L^*\!\in [1.2,\,98.8]$, $a^*\!\in [-86.2,\,98.2]$, $b^*\!\in [-107.7,\,94.5]$.
Per-language statistics appear in Tables~\ref{tab:stats_mlmc} and \ref{tab:stats_mcolornames}; examples in Table~\ref{tab:samples-combined-v2}.

\begin{table}[h]
    \centering
    \begin{minipage}[t]{0.48\textwidth}
        \footnotesize
        \centering
        \setlength{\tabcolsep}{2pt}
        \begin{tabular}{@{}lrrrr@{}}
            \toprule
            \textbf{Language} & \textbf{$N$} & \textbf{Unique} & \textbf{Vocab} & \textbf{Avg.\ len} \\
             & & \textbf{colours} & \textbf{(tokens)} & \textbf{(words)} \\
            \midrule
            Chinese (Zh) & 424 & 406 & 424 & 1.00 \\
            English (En) & 4\,677 & 4\,332 & 2\,270 & 1.96 \\
            French (Fr) & 289 & 282 & 209 & 1.69 \\
            German (De) & 341 & 331 & 293 & 1.32 \\
            Korean (Ko) & 1\,682 & 1\,674 & 1\,353 & 1.32 \\
            Polish (Pl) & 220 & 204 & 182 & 1.41 \\
            Portuguese (Pt) & 164 & 159 & 128 & 1.77 \\
            Russian (Ru) & 200 & 195 & 140 & 1.63 \\
            Spanish (Es) & 285 & 266 & 202 & 1.65 \\
            Swedish (Sv) & 239 & 234 & 223 & 1.16 \\
            \midrule
            \textbf{Total} & \textbf{8\,521} & \textbf{7\,107} & -- & -- \\
            \bottomrule
        \end{tabular}
        \vspace{4pt}
        \caption{Per-language statistics for the \textsc{MLMC} dataset.
        \emph{Unique colours}: distinct hex values;
        \emph{Vocab.}: unique tokens in colour names.}
        \label{tab:stats_mlmc}
    \end{minipage}%
\hfill
    \begin{minipage}[t]{0.48\textwidth}
        \footnotesize
        \centering
        \setlength{\tabcolsep}{2pt}
        \begin{tabular}{@{}lrrr@{}}
            \toprule
            \textbf{Language} & \textbf{$N$}$^{\ddagger}$ & \textbf{Vocab.} & \textbf{Avg.\ len} \\
             & & \textbf{(tokens)} & \textbf{(words)} \\
            \midrule
            English (En) & 10\,000 & 7\,302 & 2.71 \\
            French (Fr) & 3\,460 & 2\,726 & 2.64 \\
            German (De) & 390 & 638 & 2.39 \\
            Italian (It) & 1\,084 & 1\,259 & 2.32 \\
            Portuguese (Pt) & 2\,703 & 1\,997 & 2.63 \\
            Spanish (Es) & 3\,727 & 2\,780 & 2.55 \\
            \midrule
            \textbf{Total} & \textbf{21\,364} & -- & -- \\
            \bottomrule
        \end{tabular}
        \vspace{4pt}
        \caption{Per-language statistics for the \textsc{M-ColorNames} dataset.
        $^{\ddagger}$ All entries are unique colours (one description per colour).}
        \label{tab:stats_mcolornames}
    \end{minipage}
\end{table}

\paragraph{Multilingual ColorNames (\textsc{M-ColorNames}) dataset.}
%
To capture a more diverse range of figurative and semantically diverse color language, we build \textsc{M-ColorNames} from user-generated color names sourced from \url{https://colornames.org/}.
Compared to \textsc{MLMC}, these descriptions include more nouns and associative phrases (e.g., \textit{olive oil}, \textit{the swan}), and often omit basic color terms.
The original data does not include language labels, making language identification necessary but challenging: color names are short (a large fraction contains fewer than five tokens), borrow terms across languages, and can be ambiguous (e.g., ``beige''). We use a two-stage language detection pipeline.
First, we apply GlotLID~\citep{kargaran2023glotlid} to obtain fast initial language estimates for the full corpus.
Second, we use GPT-4o~\citep{hurst2024gpt} to refine these predictions, rebucketing examples where GlotLID's output is uncertain or inconsistent.
The final dataset covers six languages, achieves 100\% hue coverage, and spans $L^*\!\in [0.4,\,99.7]$, $a^*\!\in [-85.5,\,96.6]$, $b^*\!\in [-105.8,\,93.8]$.

\begin{table}[t]
\centering
\scriptsize
\setlength{\tabcolsep}{2.5pt}
\begin{tabular}{@{}l@{\,}c@{\,}l@{\qquad}l@{\,}c@{\,}l@{}}
\toprule
\multicolumn{6}{c}{\textbf{MLMC Dataset (pragmatic)}} \\
\midrule
\textbf{ZH} & \colorbox[RGB]{0,253,255}{\phantom{\rule{0.75em}{0.55em}}} & \begin{CJK}{UTF8}{mj}螢光藍\end{CJK} (\textit{fluorescent blue}) & \textbf{RU} & \colorbox[RGB]{105,255,0}{\phantom{\rule{0.75em}{0.55em}}} & \begin{otherlanguage}{russian}\textit{желто зеленый}\end{otherlanguage} (yellow-green) \\
 & \colorbox[RGB]{255,0,22}{\phantom{\rule{0.75em}{0.55em}}} & \begin{CJK}{UTF8}{mj}亮紅色\end{CJK} (\textit{bright red}) &  & \colorbox[RGB]{0,43,255}{\phantom{\rule{0.75em}{0.55em}}} & \begin{otherlanguage}{russian}\textit{ультрамариновый}\end{otherlanguage} (ultramarine) \\
\midrule
\textbf{KO} & \colorbox[RGB]{0,180,255}{\phantom{\rule{0.75em}{0.55em}}} & \begin{CJK}{UTF8}{mj}연한파랑\end{CJK} (\textit{light blue}) & \textbf{PL} & \colorbox[RGB]{0,200,255}{\phantom{\rule{0.75em}{0.55em}}} & \textit{niebieski jasny} (light blue) \\
 & \colorbox[RGB]{255,0,234}{\phantom{\rule{0.75em}{0.55em}}} & \begin{CJK}{UTF8}{mj}형광 핑크\end{CJK} (\textit{fluorescent pink}) &  & \colorbox[RGB]{255,0,123}{\phantom{\rule{0.75em}{0.55em}}} & \textit{różowoczerwony} (pink-red) \\
\midrule
\textbf{ES} & \colorbox[RGB]{0,255,143}{\phantom{\rule{0.75em}{0.55em}}} & \textit{verde oscuro} (dark green) & \textbf{SV} & \colorbox[RGB]{0,255,215}{\phantom{\rule{0.75em}{0.55em}}} & \textit{turkås} (turquoise) \\
 & \colorbox[RGB]{116,0,255}{\phantom{\rule{0.75em}{0.55em}}} & \textit{morada} (purple) &  & \colorbox[RGB]{216,0,255}{\phantom{\rule{0.75em}{0.55em}}} & \textit{lilla} (purple) \\
\midrule
\textbf{FR} & \colorbox[RGB]{0,136,255}{\phantom{\rule{0.75em}{0.55em}}} & \textit{ciel} (sky blue) & \textbf{PT} & \colorbox[RGB]{0,255,105}{\phantom{\rule{0.75em}{0.55em}}} & \textit{verde piscína} (pool green) \\
 & \colorbox[RGB]{255,0,167}{\phantom{\rule{0.75em}{0.55em}}} & \textit{rose chaud} (warm pink) &  & \colorbox[RGB]{0,87,255}{\phantom{\rule{0.75em}{0.55em}}} & \textit{navy} (navy blue) \\
\midrule
\textbf{DE} & \colorbox[RGB]{0,255,78}{\phantom{\rule{0.75em}{0.55em}}} & \textit{hrllgrün} (light green) & \textbf{EN} & \colorbox[RGB]{167,236,213}{\phantom{\rule{0.75em}{0.55em}}} & \textit{dull teal} \\
 & \colorbox[RGB]{107,0,255}{\phantom{\rule{0.75em}{0.55em}}} & \textit{blauviolett} (blue-violet) &  & \colorbox[RGB]{141,129,139}{\phantom{\rule{0.75em}{0.55em}}} & \textit{shark skin} \\
\midrule
\multicolumn{6}{c}{\textbf{M-ColorNames Dataset (creative)}} \\
\midrule
\textbf{EN} & \colorbox[RGB]{9,65,21}{\phantom{\rule{0.75em}{0.55em}}} & \textit{garden's darkest night} & \textbf{DE} & \colorbox[RGB]{210,217,207}{\phantom{\rule{0.75em}{0.55em}}} & \textit{seltsam gewachsene birke} (strangely grown birch) \\
 & \colorbox[RGB]{135,123,92}{\phantom{\rule{0.75em}{0.55em}}} & \textit{sand love} &  & \colorbox[RGB]{111,112,118}{\phantom{\rule{0.75em}{0.55em}}} & \textit{beim jakobsbrunnen} (at jacob's well) \\
\midrule
\textbf{ES} & \colorbox[RGB]{91,107,49}{\phantom{\rule{0.75em}{0.55em}}} & \textit{cactus cesped} (cactus lawn) & \textbf{IT} & \colorbox[RGB]{65,39,0}{\phantom{\rule{0.75em}{0.55em}}} & \textit{terra concimata} (fertilized earth) \\
 & \colorbox[RGB]{107,68,2}{\phantom{\rule{0.75em}{0.55em}}} & \textit{tierra mojada} (wet earth) &  & \colorbox[RGB]{162,159,151}{\phantom{\rule{0.75em}{0.55em}}} & \textit{necropoli di palastreto} \\
\midrule
\textbf{FR} & \colorbox[RGB]{214,48,26}{\phantom{\rule{0.75em}{0.55em}}} & \textit{petite amie} (girlfriend) & \textbf{PT} & \colorbox[RGB]{96,85,74}{\phantom{\rule{0.75em}{0.55em}}} & \textit{monte curral} (corral hill) \\
 & \colorbox[RGB]{193,191,193}{\phantom{\rule{0.75em}{0.55em}}} & \textit{le quatuor d'airain} (brass quartet) &  & \colorbox[RGB]{1,214,250}{\phantom{\rule{0.75em}{0.55em}}} & \textit{agua de beber} (drinking water) \\
\bottomrule
\end{tabular}
\vspace{4pt}
\caption{Sample color descriptions from both datasets. \textbf{Top}: MLMC (pragmatic, 2 samples $\times$ 10 languages). \textbf{Bottom}: M-ColorNames (creative, 2 samples $\times$ 6 languages). Swatches show target colours; descriptions in \textit{italics} with English translations in parentheses (omitted for English entries).}
\label{tab:samples-combined-v2}
\end{table}

\paragraph{Shared representations and descriptive statistics.}
All resources use LAB color space, which is approximately perceptually uniform. We measure spread using a dispersion score: the average $ k$-nearest-neighbor distance ($k=10$) per language.
We also compute color word proportion: the fraction of descriptions using at least one term from a language-specific color lexicon.
\textsc{MLMC} is generally more cohesive and color-term-driven, while \textsc{M-ColorNames} shows greater semantic diversity and more figurative descriptions. We use both datasets to capture this contrast.

\paragraph{\textsc{Munsell} chips (perceptual reference).}
In addition, we use \textsc{Munsell} color chips~\citep{newhall1943final} with measured CIE--LAB coordinates as perceptual reference for alignment analysis \emph{(P1)}.

\section{Experiments}
\label{sec:results}
We organize results by the two geometric properties introduced in Section~\ref{sec:problem}: \emph{(P1)} perceptual alignment and \emph{(P2)} cross-lingual symmetry. For each, we report layer-wise profiles highlighting both peak performance and final-layer behavior (relevant for downstream applications). Then, we show that spectral structure (effective rank) predicts geometric quality across both axes, validate functional consequences via task probes, and present robustness controls.

Our findings reveal a clear dissociation in how multilingual language models organize color representations: they learn a geometry that is partly \emph{universal} across languages, but not strongly \emph{perceptually grounded}. Perceptual alignment to CIE--LAB and Cross-lingual congruence follow similar behavior: scores peak in early-to-intermediate layers, then collapse toward the output ---precisely where next-token prediction occurs. However, their score ranges differ: 0.0--0.2 for \emph{P1} and 0.15--0.7 for \emph{P2}. Functionally, these geometric properties predict downstream task performance, with models achieving 96.7\% accuracy on color classification at layers where alignment peaks.

\subsection{Models and Evaluation Protocol}
\label{sec:models}
We evaluate six pretrained causal language models spanning diverse architectures, training data, and scales: Gemma-2-2B and Gemma-2-9B~\citep{team2024gemma}, Llama-3.1-8B~\citep{dubey2024llama}, Mistral-7B-v0.1~\citep{jiang2023mistral}, Qwen2.5-7B~\citep{yang2024qwen2}, and the vision-language model PaliGemma2-3B~\citep{beyer2024paligemma}. All models are text-only causal transformers except PaliGemma2, which we include to test whether visual grounding alters color geometry. Models range from 26 to 42 layers with hidden dimensions between 2\,048 and 3\,584.

For each color description, we extract hidden states at all layers (including embeddings, layer 0) and apply mean pooling over token positions to obtain sequence-level representations. We verified that mean and max pooling yield consistent results, while last-token pooling is unstable for causal models (Appendix~\ref{subsec:repr_extraction}). Unless stated otherwise, all reported metrics use mean pooling and cosine distance in representation space.

\emph{Evaluation metrics.} Perceptual alignment \emph{(P1)} is measured via centered kernel alignment (CKA) between model representations and CIE--LAB coordinates. Cross-lingual symmetry (P2) is quantified by congruence: kernel alignment between same-color representations across language pairs. Effective rank captures spectral concentration via normalized entropy of the singular value spectrum. Details in Section~\ref{sec:methods}.

\subsection{P1: Perceptual Alignment ---A Transient Window}
\label{sec:results:p1}
\begin{figure}[t]
    \centering
    \includegraphics[width=\linewidth]{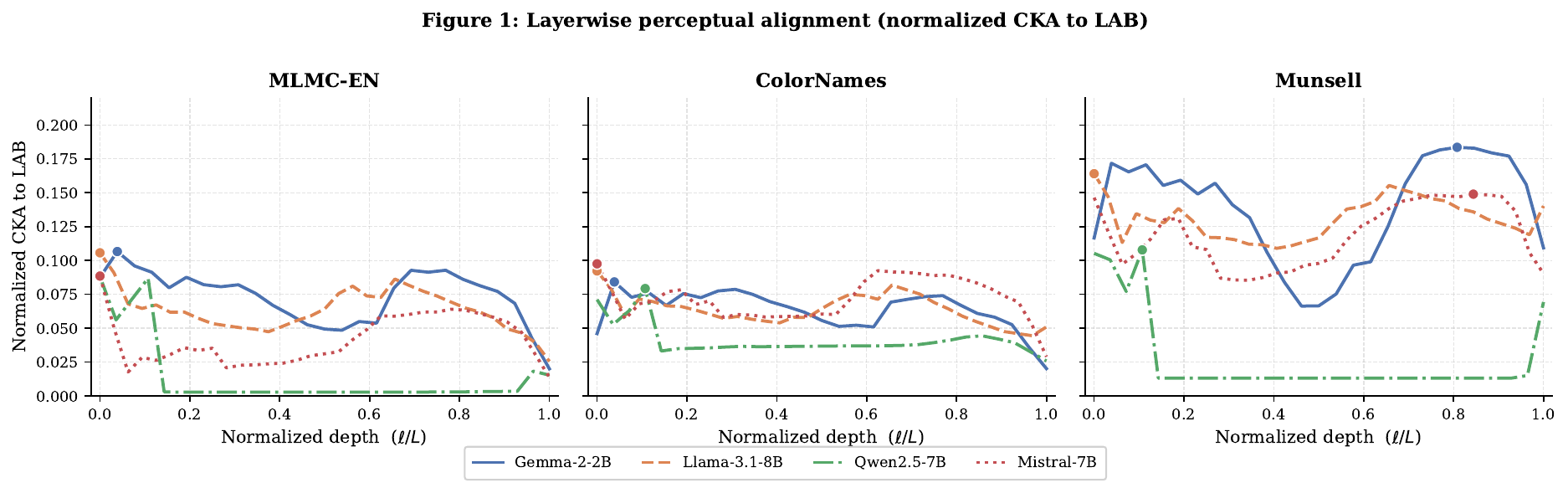}
    \caption{
    \textbf{Perceptual alignment peaks early, collapses late.}
    CKA between model representations and CIE-LAB at each layer. Filled circles mark peak per model. \textsc{MLMC-EN} and \textsc{M-ColorNames} show early-to-mid ``perceptual window'' with reduced alignment toward final layers. \textsc{Munsell} exhibits higher, more persistent alignment. Uncertainty estimates in Appendix~\ref{app:uncertainty}.
    }
    \label{fig:p1_layerwise_alignment}
\end{figure}

Language models perceptually align to CIE--LAB. However, this alignment is \emph{transient}, i.e., strongest in early-to-intermediate layers and then declining toward the output layer. Figure~\ref{fig:p1_layerwise_alignment} shows this pattern across various model families and datasets. On \textsc{MLMC-EN} and \textsc{M-ColorNames}, the alignment score peaks between layers 0--6 (depending on model depth), then it drops substantially ---by about 80\% by the final layer ---precisely where representations feed into next-token prediction. \textsc{Munsell} shows higher overall alignment and more persistent structure, though even here some models exhibit mid-layer peaks.
Alignment collapse in deeper layers can be explained by the fact that later layers optimize for language generation, i.e., lexical distinctiveness, contextual regularities, and token-level prediction. This optimization tends to discard some geometric nuance present in earlier layers. While perceptual similarity is useful for intermediate processing, it is not the primary objective of the language-modeling loss and therefore is not preserved at output.

\paragraph{Correcting the data-size bias.}
On \textsc{Munsell}, its behavior could reflect dataset-size bias ($N=122$ vs. $N>4\,000$) rather than color geometry. 
Therefore, we use matched-$n$ control to reduce this bias: \textsc{MLMC-EN} subsampled to $n=122$ five times (Table~\ref{tab:matched_n}). Results show that matching reduces the gap (\textsc{MLMC} final-layer KA increases), but the peak-then-collapse profile persists across all model families. The main difference is thus \emph{persistence across depth} (\textsc{Munsell} resists late-layer collapse better) rather than peak strength.

\begin{table}[t]
    \centering
    \small
    \begin{tabular}{@{}lrrrr@{}}
    \toprule
    Model & Peak KA ($\mu\pm\sigma$) & Final KA ($\mu\pm\sigma$) & Decay ($\mu\pm\sigma$) & Peak layer ($\mu\pm\sigma$) \\
    \midrule
    Gemma-2-2B     & 0.257 $\pm$ 0.025 & 0.091 $\pm$ 0.006 & 2.84 $\pm$ 0.28 & 1.0 $\pm$ 0.0 \\
    Gemma-2-9B     & 0.254 $\pm$ 0.028 & 0.169 $\pm$ 0.012 & 1.50 $\pm$ 0.10 & 4.4 $\pm$ 4.2 \\
    Llama-3.1-8B   & 0.240 $\pm$ 0.016 & 0.140 $\pm$ 0.021 & 1.73 $\pm$ 0.17 & 0.0 $\pm$ 0.0 \\
    Mistral-7B     & 0.242 $\pm$ 0.018 & 0.103 $\pm$ 0.015 & 2.36 $\pm$ 0.23 & 0.0 $\pm$ 0.0 \\
    Qwen2.5-7B     & 0.202 $\pm$ 0.025 & 0.079 $\pm$ 0.013 & 2.59 $\pm$ 0.30 & 0.0 $\pm$ 0.0 \\
    \bottomrule
    \end{tabular}
    \vspace{4pt}
    \caption{
    Matched-$n$ perceptual control \emph{(P1)}. \textsc{MLMC-EN} subsamples matched to \textsc{Munsell} size ($n=122$), 5 seeds.
    KA is normalized centered CKA to LAB. Matching largely reduces the apparent \textsc{Munsell} advantage while preserving the peak-to-final decay profile across model families.
    }
    \label{tab:matched_n}
\end{table}

\paragraph{Mechanism and validation: Basic Color Terms preserve perceptual structure.}
Why does some language preserve perceptual structure better? We find explicit basic color terms (BCTs) systematically enhance alignment. Splitting \textsc{MLMC-EN} by BCT presence reveals that descriptions containing BCTs (e.g., ``bright red'') yield 3--4$\times$ higher peak alignment than non-BCT descriptions (e.g., ``dull teal''), consistent across model families (Table~\ref{tab:bct_effect}). 
This mechanistic link suggests perceptual grounding emerges primarily when language explicitly evokes canonical categories. 
Figurative or indirect naming produces weaker perceptual signatures. 
We validate the evaluation pipeline using ``LAB-as-representation'' unit tests: treating LAB coordinates themselves as representations yields high alignment, while shuffled or random baselines yield near-zero alignment. 

We also verify that the near-zero kNN $\Delta E$ values sometimes observed on \textsc{Munsell} reflect the geometric density of the chip set (small nearest-neighbor LAB distances), rather than an implementation bug. These checks increase confidence that the depth profile in Figure~\ref{fig:p1_layerwise_alignment}---including the early-to-intermediate peak and reduced final-layer alignment---reflects model geometry rather than metric artifacts.

\begin{table}[t]
    \centering
    \small
    \begin{tabular}{@{}lrrrrr@{}}
    \toprule
    Model & Subset & $n$ & Peak KA & Final KA & Peak layer \\
    \midrule
    \multirow{2}{*}{Gemma-2-2B}   & BCT     & 27   & 0.4079 & 0.0934 & 1 \\
                                   & non-BCT & 27   & 0.0976 & 0.0276 & 3 \\
    \midrule
    \multirow{2}{*}{Gemma-2-9B}   & BCT     & 2\,916 & 0.4033 & 0.1603 & 1 \\
                                   & non-BCT & 1\,761 & 0.0970 & 0.0585 & 6 \\
    \midrule
    \multirow{2}{*}{Llama-3.1-8B} & BCT     & 2\,916 & 0.3555 & 0.1340 & 0 \\
                                   & non-BCT & 1\,761 & 0.0901 & 0.0538 & 0 \\
    \midrule
    \multirow{2}{*}{Mistral-7B}   & BCT     & 2\,916 & 0.3736 & 0.0711 & 0 \\
                                   & non-BCT & 1761 & 0.0897 & 0.0385 & 27 \\
    \midrule
    \multirow{2}{*}{Qwen2.5-7B}   & BCT     & 2\,916 & 0.3098 & 0.0575 & 0 \\
                                   & non-BCT & 1\,761 & 0.0720 & 0.0278 & 3 \\
    \bottomrule
    \end{tabular}
    \vspace{4pt}
    \caption{BCT-conditioned analysis on \textsc{MLMC-EN}. Descriptions containing an explicit basic color term (BCT) exhibit higher perceptual alignment (peak and final CKA) across model families, linking lexical explicitness to more perceptually structured geometry. (Gemma-2-2B rows use a smaller evaluated subset.)}
    \label{tab:bct_effect}
\end{table}

\subsection{P2: Cross-Lingual Symmetry ---Shared Early, Divergent Late}
\label{sec:results:p2}

\begin{figure}[t]
  \centering
  \includegraphics[width=\linewidth]{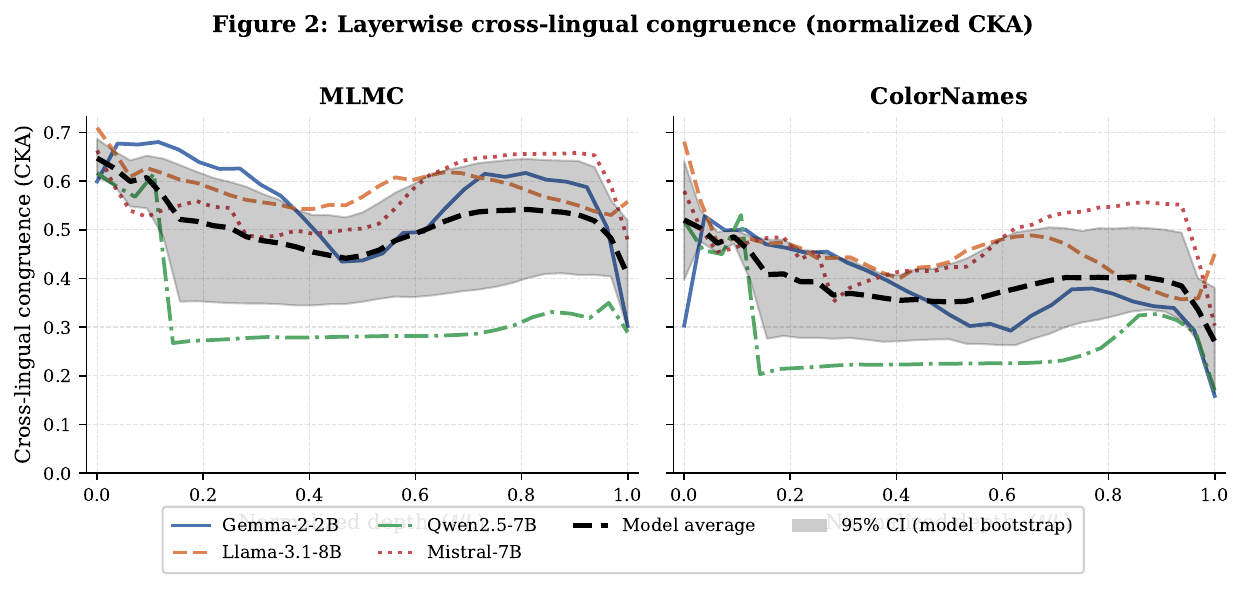}
  \caption{
  \textbf{Cross-lingual congruence degrades with depth.} 
  Mean pairwise CKA between per-language kernels on matched color sets. Thick dashed line: model average; gray band: 95\% model-bootstrap CI (robustness to model choice). High early congruence declines in later layers, with model-dependent dynamics.
  }
  \label{fig:p2_layerwise_congruence}
\end{figure}

Cross-lingual geometry is \emph{not} universally stable across depth. Instead, we find highest congruence in early layers (embeddings and shallow transformers), with model-dependent degradation as computation proceeds (Figure~\ref{fig:p2_layerwise_congruence}). At $l=0$, congruence across language pairs averages 0.61--0.71 (Table~\ref{tab:p2_congruence}), indicating substantial shared structure. By final layers, this drops to 0.29--0.56 depending on model family, with Llama/Mistral retaining more symmetry than Gemma/Qwen.

Early layers reflect shared distributional regularities: similar subword tokens, overlapping co-occurrence statistics, and universal lexical structure. Deeper layers increasingly encode \emph{language-specific} computation: tokenization idiosyncrasies, syntactic templates, and culturally contingent naming patterns. This is not collapse but specialization---later layers transform shared structure into language-conditional representations optimized for their respective linguistic contexts.
The degradation is not universal. Table~\ref{tab:p2_congruence} shows Llama-8B retains 0.56 final-layer congruence, while Gemma-2B drops to 0.30. Some models maintain relatively symmetric late-layer geometry; others exhibit sharp divergence or even mid-layer valleys followed by partial recovery. This heterogeneity suggests symmetry-breaking depends on architectural choices and training dynamics, not just depth alone.

\begin{table}[t]
      \centering
      \begin{tabular}{@{}lcrrrr@{}}
          \toprule
          \multirow{2}{*}{Model} & \multirow{2}{*}{$L$} & \multicolumn{3}{c}{Congruence} & Drift \\
          \cmidrule{3-6}
           & & $l=0$ & $l=\lceil L/2\rceil$ & $l=L$ & $l=L$ \\
          \midrule
          Gemma-2B      & 26 & 0.599 & 0.437 & 0.302 & 1.337 \\
          Gemma-9B      & 42 & 0.622 & 0.127 & 0.467 & 1.271 \\
          PaliGemma2-3B & 26 & 0.557 & 0.541 & 0.338 & 1.296 \\
          Llama-8B      & 32 & 0.710 & 0.568 & 0.558 & 1.032 \\
          Mistral-7B    & 32 & 0.663 & 0.503 & 0.479 & 1.085 \\
          Qwen-7B       & 28 & 0.617 & 0.281 & 0.289 & 1.152 \\
          \bottomrule
      \end{tabular}
      \vspace{4pt}
      \caption{Cross-lingual congruence across layers (matched-set kernels).}
      \label{tab:p2_congruence}
  \end{table}

\subsection{What Enables Geometric Preservation? Spectral Structure}
\label{sec:results:spectral}

Why do some layers preserve color geometry better than others? We find \emph{spectral structure}---specifically, effective rank---strongly predicts both perceptual alignment and cross-lingual congruence. Effective rank (ER) measures whether singular values of the representation matrix are spread broadly (high ER, many active dimensions) or concentrated in a few modes (low ER, spectral collapse).

Table~\ref{tab:unified_corr} shows correlations between layer-wise ER and both \emph{P1} alignment and \emph{P2} congruence. Across datasets and models, correlations are consistently high: $r \in [0.76, 0.99]$ for most model-dataset pairs. Layers with broad spectral support (high ER) preserve richer relational structure; layers undergoing spectral concentration (low ER) lose both alignment and symmetry.
%
This provides a unifying explanation for both the perceptual window and cross-lingual degradation. Intermediate layers maintain broad spectra, enabling complex similarity structures. Final layers undergo \emph{spectral compression}---a form of lossy dimensionality reduction optimizing for next-token prediction. This compression necessarily sacrifices geometric nuance, affecting both perceptual alignment and language-invariant structure. Effective rank thus serves as a practical diagnostic: to find layers that preserve color geometry, seek high ER.

\begin{table}[t]
    \centering
    \small
    \begin{tabular}{@{}lrrrrrr@{}}
        \toprule
        Dataset & Gemma-2B & Gemma-9B & PaliGemma2-3B & Llama-8B & Mistral-7B & Qwen-7B \\
        \midrule
        \textsc{MLMC} & 0.876 & 0.930 & 0.889 & 0.781 & 0.941 & 0.986 \\
        \textsc{M-ColorNames} & 0.820 & 0.929 & 0.907 & 0.761 & 0.755 & 0.859 \\
        \textsc{Munsell} & 0.942 & 0.980 & 0.935 & 0.851 & 0.943 & 0.996 \\
        \bottomrule
    \end{tabular}
    \vspace{4pt}
    \caption{Correlation between effective rank and key \emph{P1 \& P2} metrics (reported in the analysis).}
    \label{tab:unified_corr}
\end{table}

\subsection{Functional Consequences: Geometry Predicts Task Performance}
\label{sec:results:functional}
\begin{figure}[t]
  \centering
  \includegraphics[width=\linewidth]{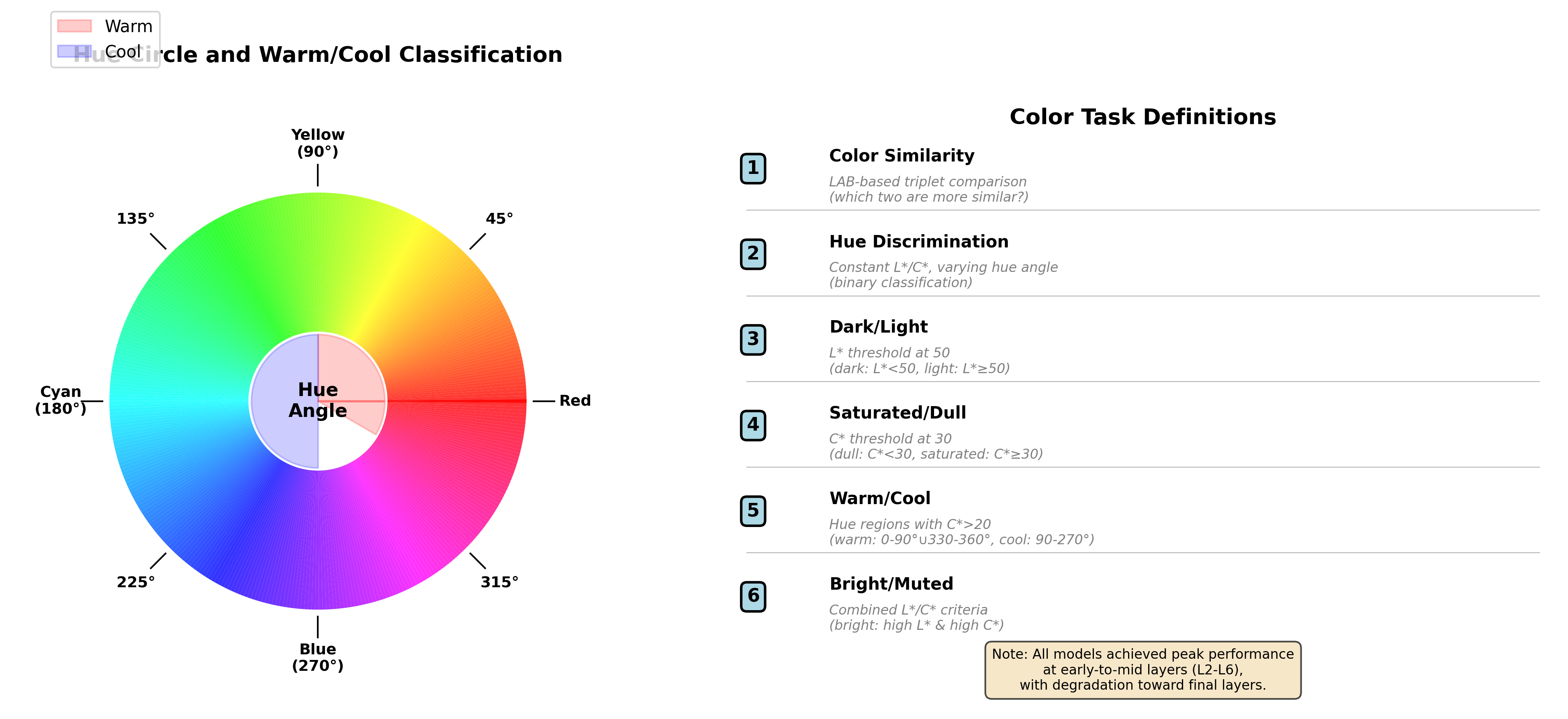}
  \caption{
  \textbf{Color task definitions.}
  \textbf{Left:} Hue circle showing the warm/cool classification task. Warm colors (red-yellow, hue angles 0--90° and 330--360° with $C^* > 20$) are highlighted in red; cool colors (yellow-green-blue-purple, 90--270°) are highlighted in blue. This task directly tests circular hue structure.
  \textbf{Right:} Summary of all six color tasks used in functional validation. Tasks probe different aspects of color space: LAB-based similarity (Task 1), hue discrimination (Task 2), lightness (Task 3), chroma (Task 4), hue regions (Task 5), and combined attributes (Task 6). All models achieved peak performance at early-to-mid layers ($l \in [2, \ldots, 6]$), with degradation toward final layers, mirroring the \emph{P1} alignment profile.
  }
  \label{fig:hue_wheel_tasks}
\end{figure}

\begin{figure}[t]
  \centering
  \includegraphics[width=0.48\linewidth]{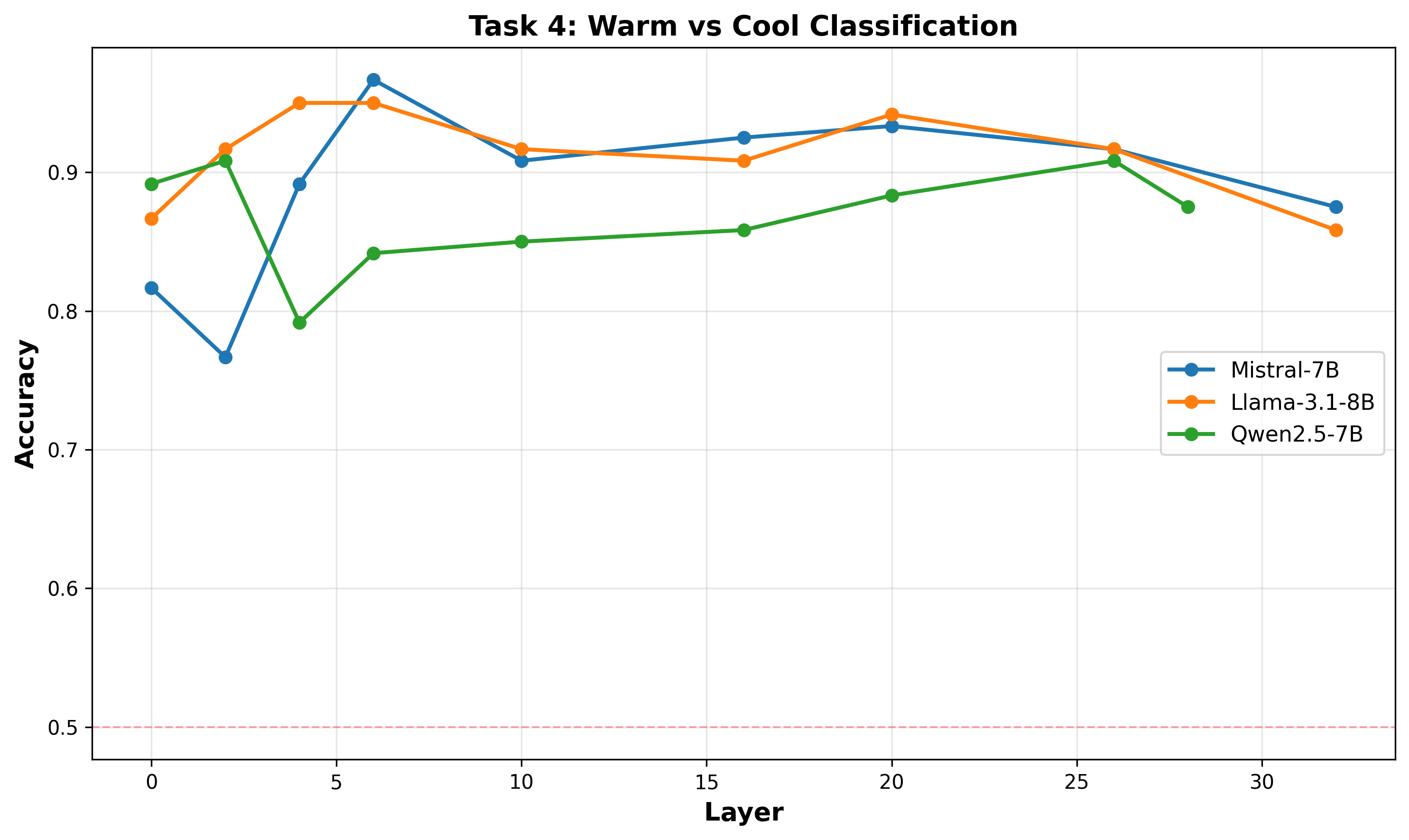}
  \hfill
  \includegraphics[width=0.48\linewidth]{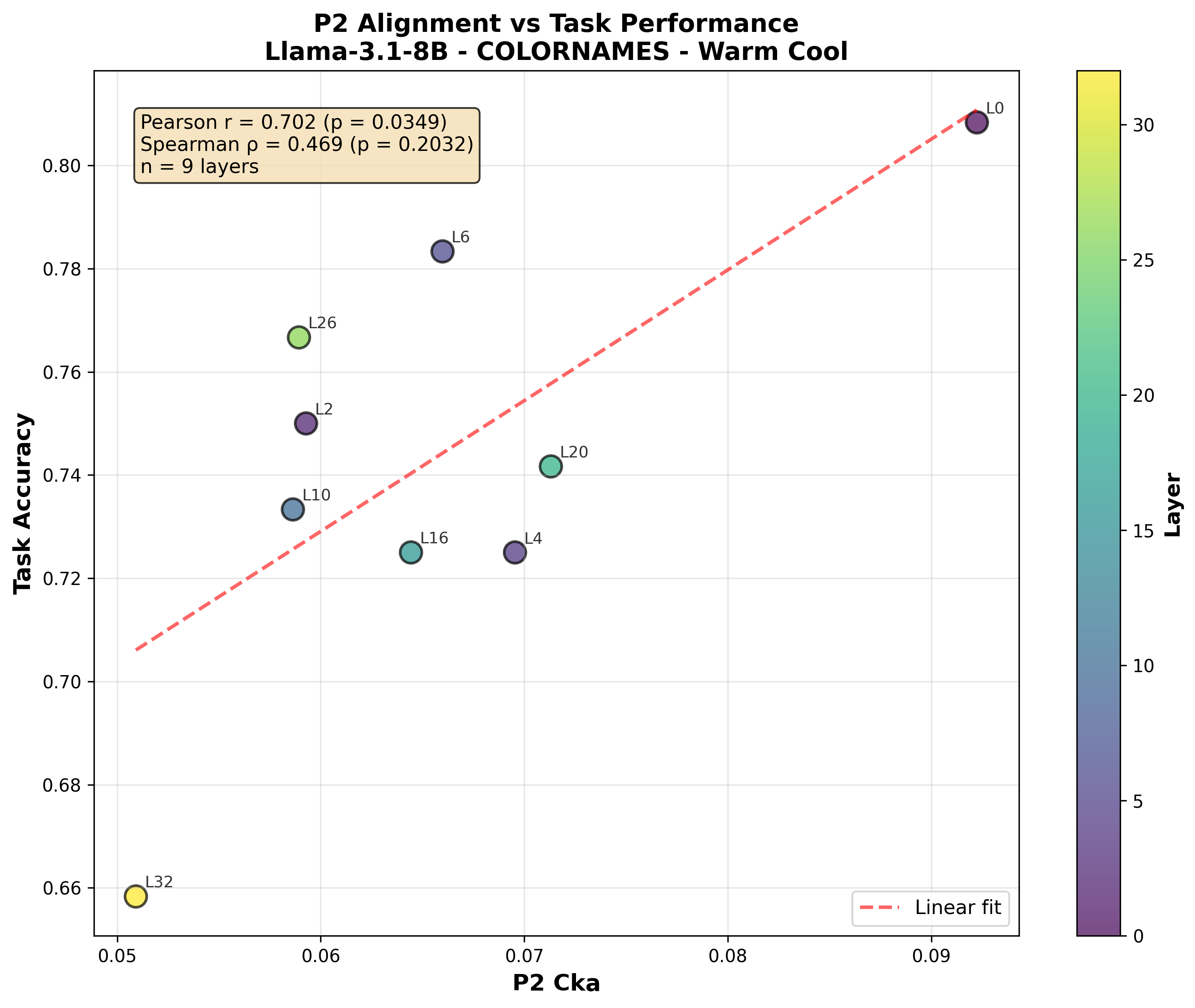}
  \caption{
  \textbf{Functional validation of perceptual alignment (P1).}
  \textbf{Left:} Layer-wise warm/cool classification accuracy for three models on MLMC (8\,521 samples). 
  \textbf{Right:} Correlation between P1 perceptual alignment (CKA) and warm/cool task accuracy across layers for Llama-3.1-8B on M-ColorNames (21\,364 samples). Each point represents one layer, colored by depth. Linear fit shows significant positive correlation ($r=0.70$, $p=0.035$), demonstrating that \emph{P1} alignment predicts functional competence: layers with stronger perceptual grounding excel at color attribute tasks. The dashed red line indicates chance performance (50\%).
  }
  \label{fig:functional_validation}
\end{figure}

Does the geometric structure we measure have functional consequences? Yes. We trained linear probes~\citep{alain2016understanding} on six color tasks (hue discrimination, warm/cool classification, bright/muted, etc.) at different layers and find that \emph{task performance tracks perceptual alignment} (see Figure~\ref{fig:hue_wheel_tasks}). Models achieve highest accuracy exactly where \emph{P1} alignment peaks, then degrade toward final layers---mirroring the geometric collapse.
On warm/cool classification (testing hue structure), Mistral-7B reached \textbf{96.7\% accuracy at $l=6$} on \textsc{MLMC}---exactly where perceptual alignment peaks (Figure~\ref{fig:functional_validation}). Llama-3.1-8B achieved 95.0\% at $l=4$, Qwen2.5-7B 90.8\% at $l=2$. Performance then degrades toward final layers, mirroring geometric collapse. This pattern generalizes: all six color tasks (color similarity, hue discrimination, brightness, saturation, warm/cool, and bright/muted) peak at layers 2--10, then decline.

Layer-wise CKA directly predicts task accuracy. For Llama-3.1-8B on \textsc{M-ColorNames}, correlation between CKA and warm/cool accuracy is $r=0.70$ ($p=0.035$). Distance-based metrics show even stronger correlations: Spearman $\rho$ up to 0.96, neighborhood preservation up to 0.86. Across 120 correlations (4 metrics $\times$ 5 tasks $\times$ 6 model-dataset pairs), 12.5\% reach significance with mean $|r|=0.38$. This confirms perceptual alignment is not just descriptive but \emph{functionally predictive}: layers preserving LAB structure excel at color tasks.


For color-related downstream applications, extract representations from early-to-mid layers (where alignment peaks), not final layers, in line with evidence that intermediate layers can outperform the last layer on general downstream tasks~\citep{skean2025layer}. This has immediate practical value: choosing the right layer can yield 2--3$\times$ improvement in probe accuracy.


\section{Discussion}
\label{sec:discussion}

\subsection{The Dissociation: Universal Geometry Without Perceptual Grounding}

We find that multilingual language models learn color representations that exhibit a striking dissociation: the geometry is partly \emph{universal} (congruent across languages in early-to-mid layers) but not strongly \emph{perceptually grounded} (weak alignment to CIE--LAB, transient across depth). This challenges the notion that universal representations in large-scale language models necessarily reflect perceptual or world-grounded structure.

The dissociation is most evident at the output layer, where next-token prediction occurs. Here, cross-lingual symmetry persists at moderate-to-high levels for many models, yet perceptual alignment collapses. This suggests that models learn a color-language geometry optimized for \emph{linguistic categorization}---supporting lexical distinctions and pragmatic naming---rather than preserving perceptual similarity. The geometry reflects distributional regularities in how languages name colors, not the perceptual organization of color space itself.

This finding refines recent claims about ``platonic representations'' in large-scale models. While some domains may converge toward perceptually or physically grounded structure (e.g., object shape, spatial relations), color reveals a domain where convergence occurs along a \emph{linguistic} axis. The universality we observe is not perceptual universality but \emph{distributional universality}: patterns shared across languages because of how color terms co-occur with other linguistic elements in text.

\subsection{Why Does the Perceptual Window Exist and Collapse?}

The transient perceptual window---strongest alignment in early-to-mid layers, collapsing toward the output---suggests a specific computational trajectory. Early layers aggregate subword and lexical information, where color terms co-occur with perceptually correlated contexts (e.g., ``sky blue'', ``blood red''). This incidental correlation yields moderate perceptual alignment without explicit perceptual supervision.

As depth increases, representations are transformed toward features directly useful for next-token prediction. Late layers prioritize syntactic, pragmatic, and distributional structure over perceptual similarity. The BCT effect reinforces this: descriptions using basic color terms (e.g., ``dark green'') preserve more perceptual structure than figurative or associative descriptions (e.g., ``garden's darkest night''), because BCTs anchor explicit lexical categories that correlate with perceptual regions.

Why does alignment collapse at the output rather than plateau? One explanation is \emph{spectral compression}: as effective rank drops in late layers, the representational capacity to preserve fine-grained perceptual distinctions degrades. Another is \emph{objective mismatch}: next-token prediction rewards linguistic coherence and pragmatic plausibility, not perceptual fidelity. Models may actively overwrite perceptual correlates with linguistic structure because the latter is more task-relevant.

The matched-$N$ controls show this is not a sample-size artifact: even when \textsc{MLMC} is downsampled to match \textsc{Munsell}'s size, the peak-then-collapse profile persists. This indicates the perceptual window reflects model computation, not dataset density.

\subsection{Why Does Cross-Lingual Symmetry Degrade?}

Cross-lingual congruence is substantial at embeddings, reflecting shared subword statistics and co-occurrence structure across languages. As depth increases, congruence often declines, with model-dependent trajectories: some architectures retain high symmetry deeper, others show sharper degradation or late-layer recovery.

This evolution likely reflects a transition from \emph{distributional universality} in early layers to \emph{language-specific functional specialization} in late layers. Early layers aggregate evidence from shared multilingual pretraining, where cross-lingual transfer is strong. Deeper layers increasingly encode language-conditional transformations: syntactic ordering, morphological patterns, culturally specific naming conventions, and tokenization idiosyncrasies. The fact that some models retain higher late-layer congruence suggests architectural or objective factors (e.g., parameter sharing, cross-lingual alignment during training) can mitigate this specialization.

Spectral structure again provides a mechanistic link: layers undergoing compression (low effective rank) lose both perceptual alignment \emph{and} cross-lingual symmetry. This suggests a common underlying mechanism: as representational capacity contracts, models sacrifice fine-grained distinctions---both perceptual and cross-lingual---in favor of coarser linguistic categories optimized for generation.

\subsection{Implications for Representation Learning}

\paragraph{Universal does not mean grounded.}
These findings establish a boundary condition on claims about ``platonic'' or perceptually grounded representations in large-scale language models. Universal structure can emerge without grounding: models converge on shared color-language geometry not because they learn perceptual similarity, but because they learn shared distributional structure in how languages name colors. This suggests that universality in LMs should be evaluated against \emph{multiple reference templates}---perceptual, lexical, functional---rather than assumed to reflect world structure by default.

\paragraph{Layer selection matters.}
The transient perceptual window has practical implications. Tasks requiring perceptual reasoning (e.g., color-based visual QA, accessibility tools) may benefit from probing early-to-mid layers rather than final representations. Conversely, linguistic generation tasks (naming, pragmatic reasoning) align with late-layer representations optimized for next-token prediction. This suggests layer-selective probing strategies should match the task's demands.

\paragraph{Spectral structure as a geometric diagnostic.}
Effective rank provides a simple, interpretable signal for when geometric quality degrades. Layers undergoing spectral compression lose both perceptual alignment and cross-lingual symmetry. This offers a practical tool for model developers: monitoring spectral concentration can flag when representational capacity contracts, potentially guiding interventions (e.g., regularization, architectural choices) to preserve geometric fidelity.

\paragraph{Implications for multimodal models.}
While our focus is text-only models, the dissociation we observe motivates questions for vision-language architectures. Do models with visual grounding exhibit stronger perceptual alignment at the output? Or does the linguistic objective still dominate late layers? Preliminary evidence (e.g., deviations in multimodal models noted in §\ref{sec:results}) suggests that injecting visual supervision can reweight which factors are preserved, but the extent to which linguistic objectives override perceptual structure remains an open question.

\subsection{Limitations and Future Directions}

\paragraph{Dataset scope and sampling.}
Our datasets differ in size and density (\textsc{Munsell}: $n=122$ chips; \textsc{MLMC}: $\sim$8.5k descriptions; \textsc{M-ColorNames}: $\sim$21k). While matched-$N$ controls verify that qualitative trends persist under matched evaluation, absolute alignment magnitudes can depend on sampling density. Future work could extend this to systematically sampled color spaces (e.g., CIE--LAB grids) to isolate density effects more precisely.

\paragraph{Methodological choices.}
We use CKA with RBF kernels as our primary alignment metric. While robust to orthogonal transformations and scaling, absolute values depend on the kernel bandwidth, which we set with the median heuristic; we do not evaluate other bandwidth choices. Alternative similarity metrics (e.g., mutual information, Procrustes distance) could also provide complementary perspectives. Pooling choices also matter: mean/max pooling is stable across models, but last-token pooling shows model-dependent instability (Table~\ref{tab:pooling_robust}). We recommend content-aggregating pooling for causal LM geometry probes.

\paragraph{Reliability of CKA.}
Prior work has shown that CKA is not a fully reliable similarity measure: it can
disagree with other measures, and it can change under transformations that leave
a model's behavior unchanged~\citep{ding2021grounding,davari2023reliability}.
These studies compare neural networks with each other, whereas we compare a network to a fixed perceptual reference (CIE--LAB), so we do not claim that their findings carry over directly. To limit our reliance on any single measure, we also report Spearman correlation, matched-$N$ controls, and sanity-check baselines, and we relate alignment to probe accuracy. We interpret CKA values as relative comparisons across layers, not as absolute measures of perceptual
grounding.

\paragraph{Beyond color.}
Color is a controlled, well-studied domain with established perceptual metrics. Whether the dissociation we observe generalizes to other perceptual domains (e.g., shape, spatial relations, auditory concepts) remains open. Some domains may show stronger perceptual grounding (e.g., spatial prepositions), others may exhibit similar distributional convergence. Systematic cross-domain studies would clarify when and why LMs preserve perceptual structure.

\paragraph{Mechanistic interpretability.}
We identify \emph{where} (early-to-mid layers) and \emph{when} (spectral compression) geometry degrades, but not precisely \emph{how}. Future work using causal intervention methods (e.g., activation patching, ablations) could trace which computations overwrite perceptual structure and which preserve cross-lingual symmetry. This would move from correlation to causation, clarifying the mechanisms underlying the dissociation.

\section{Conclusions}

Multilingual language models learn color representations that are partly universal but not perceptually grounded. The geometry reflects distributional regularities in how languages name colors, not the perceptual organization of color space. This dissociation is clearest at the output layer, where next-token prediction occurs: cross-lingual symmetry persists, but perceptual alignment collapses. A transient perceptual window exists in early-to-mid layers, but late layers prioritize linguistic structure over perceptual fidelity. Spectral compression explains when geometry degrades: layers with low effective rank lose both alignment and symmetry. Functionally, geometric quality predicts task performance, with practical implications for layer selection and probing strategies. These findings challenge the notion that scale and multilingual training automatically yield perceptually grounded representations, suggesting instead that universality in LMs reflects shared linguistic structure, not necessarily world structure.

\section*{AI usage disclosure}
Large language model tools were used during research development for language editing, code
assistance and debugging, literature discovery, and workflow support. 
The authors determined the research questions and experimental protocols, inspected and verified experimental outputs, and
remain responsible for all claims, analyses, and submitted content.

\bibliographystyle{unsrtnat}
\bibliography{references}

\appendix 
\label{sec:appendix}

\section{Representation extraction and pooling}
\label{subsec:repr_extraction}

Let $\mathcal{D}=\{(\mathbf{c}_i,\, \mathbf{x}_i)\}_{i=1}^{N}$ denote a dataset of color chips $\mathbf{c}_i$ with associated natural-language description $\mathbf{x}_i$ (possibly multilingual). We tokenize $\mathbf{x}_i$ with the model's tokenizer and run a forward pass to obtain hidden states at every layer $l\in\{0,\ldots,L\}$ (including the embedding layer $l=0$). We write the sequence of hidden states as
\begin{equation}
  \mathbf{H}_i^{(l)} \in \mathbb{R}^{T_i\times d_l},
\end{equation}
where $T_i$ is the sequence length and $d_\ell$ is the hidden size at layer $\ell$.

We map each sequence to a single vector via a pooling operator $\pi$:
\begin{equation}
  \mathbf{z}_i^{(l)} \;=\; \pi\!\left(\mathbf{H}_i^{(l)}\right) \in \mathbb{R}^{d_l},
\end{equation}
and stack vectors row-wise to obtain
\begin{equation}
  \mathbf{Z}^{(l)} \;=\; \begin{bmatrix} (\mathbf{z}_1^{(l)})^\top \\ \vdots \\ (\mathbf{z}_N^{(l)})^\top \end{bmatrix} \in \mathbb{R}^{N\times d_l}.
\end{equation}
Unless stated otherwise, we use a content-aggregating pooling strategy (mean or max pooling over the token dimension, excluding padding). We report ablations for alternative pooling choices (e.g., last-token, [CLS] when available). In our experiments, mean and max pooling behave similarly across model families, whereas last-token pooling can be unstable and model-dependent for causal LMs (Table~\ref{tab:pooling_robust}).

\section{Basic Color Term Analysis}
\label{app:bct}
To characterize the linguistic content of color descriptions across languages, we analyze the usage of \emph{Basic Color Terms} (BCTs) following \citet{berlin1991basic}. BCTs are psychologically salient, monolexemic color words that are widely used and relatively stable across speakers (e.g., ``red'', ``blue'', ``green'').

\paragraph{Dictionaries.}
For each language, we construct a BCT dictionary covering the 11 canonical categories (white, black, red, green, yellow, blue, brown, purple, pink, orange, gray). For two languages we incorporate linguistically motivated extensions:

\textbf{Korean (31 terms).}
Modern Korean exhibits multiple co-existing color-naming strata, including (i) native Korean forms, (ii) adjective forms, (iii) Sino-Korean terms, and (iv) loanwords; these correlate with register (informal/formal/technical/trendy) and are all productive in contemporary usage \citep{sohn2001korean, lee2001korean}. We include representative terms from all strata, e.g., native/adjectival \begin{CJK}{UTF8}{mj}빨강/빨간\end{CJK} (``red''), Sino-Korean \begin{CJK}{UTF8}{mj}적색\end{CJK} (``red''), and loanword \begin{CJK}{UTF8}{mj}레드\end{CJK} (``red'').

\textbf{Russian (12 terms).}
Russian distinguishes \emph{siniy} (dark blue) and \emph{goluboi} (light blue), a categorical split absent in many other languages. Prior work shows categorical perception effects at the \emph{siniy--goluboi} boundary comparable to cross-category boundaries elsewhere, suggesting both are psychologically basic \citep{winawer2007russian}. We therefore include both in the Russian dictionary.

\paragraph{Matching procedure.}
We mark a description as containing a BCT if it includes any dictionary term under case-folding. To reduce spurious substring matches (e.g., accidental matches within longer words), we implement boundary-aware matching (whitespace/punctuation boundaries for space-delimited scripts; language-specific tokenization where applicable).

\section{Uncertainty}
\label{app:uncertainty}
\subsection{Metric validation and implementation checks}
\label{subsec:metric_validation}

Because several metrics rely on pairwise distances and nearest-neighbor queries, we apply a small set of implementation checks to ensure comparability across datasets and models. In particular, we (i) use a consistent $\Delta E$ definition in LAB throughout, (ii) exclude self-matches when computing kNN-based quantities, (iii) confirm that cosine distance and $\ell_2$ distance on unit-normalized vectors produce equivalent neighbor orderings up to monotone transformations, and (iv) check trivial baselines for which the expected outcome is known (e.g., using LAB coordinates as representations should yield near-perfect alignment). These checks are intended to prevent spurious discontinuities due to unit or indexing errors.

\subsection{Uncertainty estimates and statistical protocol}
\label{subsec:uncertainty}

\paragraph{Uncertainty.}
All reported quantities can be paired with bootstrap confidence intervals over
samples (and over prompts/pooling seeds when applicable), which we use to assess
stability across datasets, layers, and models.

Unless otherwise stated, we estimate uncertainty via nonparametric bootstrap over chips.
For each metric $m$ at layer $\ell$, we resample the $N$ chips with replacement $B$ times (default $B=500$) and recompute $m$.
We report the bootstrap mean and 95\% confidence intervals (2.5/97.5 percentiles).
For multilingual comparisons, bootstraps preserve the matched-set structure by resampling matched chip indices.

\paragraph{Model-bootstrap bands.}
When aggregating trends across a small set of model families (e.g., the gray band in Figure~\ref{fig:p2_layerwise_congruence}), we use a \emph{model bootstrap} that resamples model curves rather than chips. This band reflects robustness to model choice and should not be interpreted as sample-level statistical significance.

\subsection{Controls and ablations}
\label{subsec:controls}

We include the following controls to stress-test conclusions:
\begin{itemize}
  \item \textbf{Matched-$N$ perceptual control.} We downsample large datasets (e.g., \textsc{MLMC}) to the \textsc{Munsell} size $n=|\mathcal{C}_{\text{\textsc{Munsell}}}|$ (and repeat across seeds) to disentangle dataset scale from perceptual alignment.
  \item \textbf{Distance choice.} On unit-normalized vectors, cosine and $\ell_2$ distance induce equivalent neighbor orderings (Appendix~\ref{subsec:metric_validation}); all reported results use cosine distance.
  \item \textbf{Pooling choice.} Mean pooling vs max pooling vs last-token / [CLS] pooling when available.
\end{itemize}

\paragraph{Pooling robustness is family-dependent; mean/max are stable defaults.}
We evaluate three pooling strategies for causal LMs (max, mean, last-token) on \textsc{MLMC-EN} ($N\le 1000$). As shown in Table~\ref{tab:pooling_robust}, mean and max pooling yield broadly similar peak locations and peak-to-final decay across all tested families, indicating that the main \emph{P1} pattern is not sensitive to this choice of content-aggregating pooling. In contrast, last-token pooling is unstable and model-dependent: it collapses alignment for Gemma and Mistral, but can yield higher peaks for Llama and Qwen. This suggests that sequence-end representations can be disproportionately shaped by token-position or prompt-format effects, making last-token pooling a poor default for geometry probes of causal LMs. We therefore use mean/max pooling as the default and treat last-token pooling as a non-robust ablation.

\begin{table}[t]
\centering
\small
\begin{tabular}{@{}lrrrrr@{}}
\toprule
Model & Pooling & Peak KA & Peak layer & Final KA & Decay \\
\midrule
\multirow{3}{*}{Gemma-2-2B}     & max  & 0.2025 & 1 & 0.0500 & 4.05$\times$ \\
     & mean & 0.2041 & 1 & 0.0422 & 4.84$\times$ \\
     & last & 0.0301 & 1 & 0.0121 & 2.50$\times$ \\
\midrule
\multirow{3}{*}{Gemma-2-9B}     & max  & 0.2014 & 2 & 0.0946 & 2.13$\times$ \\
    & mean & 0.2154 & 1 & 0.0950 & 2.27$\times$ \\
    & last & 0.0784 & 1 & 0.0392 & 2.00$\times$ \\
\midrule
\multirow{3}{*}{Llama-3.1-8B}   & max  & 0.1954 & 0 & 0.0722 & 2.71$\times$ \\
   & mean & 0.2109 & 0 & 0.0707 & 2.98$\times$ \\
   & last & 0.2370 & 3 & 0.0934 & 2.54$\times$ \\
\midrule
\multirow{3}{*}{Mistral-7B}     & max  & 0.2056 & 0 & 0.0462 & 4.45$\times$ \\
     & mean & 0.2233 & 0 & 0.0459 & 4.87$\times$ \\
     & last & 0.0379 & 0 & 0.0173 & 2.19$\times$ \\
\midrule
\multirow{3}{*}{Qwen2.5-7B}     & max  & 0.1522 & 0 & 0.0270 & 5.65$\times$ \\
     & mean & 0.1948 & 0 & 0.0243 & 8.01$\times$ \\
     & last & 0.2340 & 0 & 0.0615 & 3.80$\times$ \\
\bottomrule
\end{tabular}
\vspace{4pt}
\caption{Pooling robustness \emph{(P1)}. Mean and max pooling give broadly consistent peak and decay across families, while last-token pooling is unstable (collapsing alignment for Gemma/Mistral but yielding higher peaks for Llama/Qwen). We use mean/max pooling as the default and treat last-token as an ablation.}
\label{tab:pooling_robust}
\end{table}

\end{document}